\documentclass[runningheads]{llncs}
\pdfoutput=1

\usepackage[utf8]{inputenc}
\usepackage{graphicx}
\usepackage{enumitem}
\usepackage{amsmath}
\usepackage{empheq}
\usepackage{amsfonts}
\usepackage{amssymb}
\usepackage{tikz}
\usepackage{physics}
\usepackage{rsfso}
\usepackage{algorithm}
\usepackage{algpseudocode}
\usepackage{bbold}
\usepackage{multirow}
\usepackage{float}
\usepackage{xcolor}
\usepackage{hf-tikz}
\usepackage{epstopdf}
\usetikzlibrary{automata,positioning,shapes}

\begin{document}

\title{Neural Networks Reduction via Lumping}
\titlerunning{NNs Reduction via Lumping}

\author{Dalila Ressi\inst{1}\orcidID{0000-0001-5291-5438} \and  Riccardo Romanello \inst{1}\orcidID{0000-0002-2855-1221} \and Carla Piazza\inst{1}\orcidID{0000-0002-2072-1628} \and Sabina Rossi\inst{2}\orcidID{0000-0002-1189-4439}}



\institute{
  Universit\`a di Udine, Italy\\
  \email{\{dalila.ressi, riccardo.romanello, carla.piazza\}@uniud.it}
  \and
Universit\`a Ca' Foscari Venezia, Italy\\
 \email{sabina.rossi@unive.it}
}


\maketitle

\begin{abstract} 
The increasing size of recently proposed Neural Networks makes it hard to implement them on embedded devices, where memory, battery and computational power are a non-trivial bottleneck. For this reason during the last years network compression literature has been thriving and a large number of solutions has been been published to reduce both the number of operations and the parameters involved with the models. Unfortunately, most of these reducing techniques are actually heuristic methods and usually require at least one re-training step to recover the accuracy. 

The need of procedures for model reduction is well-known also in the fields of Verification and Performances Evaluation, where large efforts have been devoted to the definition of quotients that preserve the observable underlying behaviour.

In this paper we try to bridge the gap between the most popular and very effective network reduction strategies and formal notions, such as lumpability, introduced for verification and evaluation of Markov Chains. Elaborating on lumpability we propose a pruning approach that reduces the number of neurons in a network without using any data or fine-tuning, while completely preserving the exact behaviour. 
Relaxing the constraints on the exact definition of the quotienting method we can give a formal explanation of some of the most common reduction techniques.

\end{abstract}

\section{Introduction}
Since 2012, when AlexNet \cite{krizhevsky2012imagenet} won the famous ImageNet Large Scale Visual Recognition Challenge (ILSVRC), the number of proposed \textit{Artificial Neural Network} (\textit{ANN} or \textit{NN}) architectures has increased exponentially. Their intrinsic flexibility, together with the superior performance they can achieve, made neural networks the tool of choice to solve a wide variety of tasks. 
As these models have evolved to process large amount of data or to solve complicated tasks, their complexity has also increased at same pace~\cite{deng2020model}. Such elaborate and deep networks are the foundation of \textit{Deep Learning} (\textit{DL}) and they stand out both for the large number of layers they are made of and for the higher level of accuracy they can reach on difficult tasks \cite{xiao2018dynamical}.

While the academic community mostly focused their efforts in training large and deep models~\cite{kolesnikov2020big,dai2021coatnet,yu2022coca}, being able to adopt such networks in embedded devices resulted to be a problem. Physical constraints such as battery, memory and computational power greatly limit both the number of parameters used to the define the architecture and the number of Floating Point Operations (FLOPs) required to be computed at inference time.
%
A commonly used strategy to address this problem is called \textit{Network Compression}. Compression literature has had a substantial growth during the last years, and for this reason there are many different ways to group together methods reducing a model in similar~ways. 

Methods focusing on finding the best possible structure to solve a particular tasks can be grouped together as \textit{Architecture-related} strategies. These kind of methods usually require to train the network from scratch each time the structure is modified. In particular, \textit{Neural Architecture Search (NAS)} techniques aim to find the best possible architecture for a certain task with minimal human intervention \cite{ren2021comprehensive,elsken2019neural,liu2021survey}. This is usually made possible by modelling the search as an optimization problem and applying \textit{Reinforcement Learning (LR)}-based methods to find the best architecture \cite{zoph2016neural,baker2016designing}. 
In this group we can also find \textit{Tensor Decomposition}, where matrix decomposition/factorization principles are applied to the $d$-dimensional tensors in neural networks. Tensor decomposition generalizes the widely used Principal Component Analysis (PCA) and Singular Value Decomposition (SVD) to an arbitrary number of dimensions \cite{carroll1970analysis,harshman1970foundations,tucker1966some}. The goal of these techniques is to reduce the rank of tensors in order to efficiently decompose them into smaller ones and drastically reduce the number of operations \cite{deng2020model}. As the rank of a tensor is usually far from being small, the most common solutions are to either to force the network to learn filters with small rank either to use an approximated decomposition \cite{denton2014exploiting}.

Using a similar approach \textit{Lightweight} or \textit{Compact Networks} focus on modifying the design of the architecture such that it performs less operations while maintaining the same capability. It is the case of the MobileNet series \cite{howard2017mobilenets,sandler2018mobilenetv2,howard2019searching}, ShuffleNet series \cite{zhang2018shufflenet,ma2018shufflenet}, and EfficientNet series \cite{tan2019efficientnet,tan2021efficientnetv2}. They exploit the idea of using 1$\times$1 filters introduced by Network in Network \cite{lin2013network} and GoogLeNet \cite{szegedy2015going,szegedy2016rethinking} in their inception modules. A similar concept is explored by the SqueezeNet \cite{iandola2016squeezenet} architecture in their \textit{Fire module}, where they substitute the classical convolutional layers such that they can achieve the same accuracy of AlexNet on ImageNet dataset but with a model 510 times smaller.

A different methodology consists in training a big model from the start, and then \textit{Pruning} superfluous parameters. 
In particular, \textit{Weight Pruning} consists in zeroing connections or parameters already close to zero \cite{lecun1990optimal}, but more elaborated methods can also take into consideration the impact of the single weights on the final results \cite{han2015learning}. Even if weight pruning is a very powerful tool to reduce the network parameters~\cite{frankle2018lottery}, its major drawback is that it does not actually reduce the number of FLOPs at inference time.

A more effective solution consists instead in skipping completely some of the operations. It is the case of \textit{Filter Pruning}, where whole nodes or filters (in case of convolutional layers) are removed from the architecture. Pruning usually requires some degree of re-training to recover the lost accuracy due to the reduced network capability, but an interesting phenomena that happens in the early stages of pruning is that most of the times the test accuracy actually increases, due to the regularization effect that pruning unnecessary parameters has on the network. While weight pruning allows more control on what parameters to remove, filter pruning is usually the best solution compression-wise as it allows to drastically reduce the network parameters such that the models can be actually implemented in small embedded devices \cite{ressi2022relevance}. 

Another technique often used in conjunction with pruning is called \textit{quantization} \cite{han2015deep}. While pruning aims to reduce the number of parameters, quantization instead targets their precision. As the weights are usually represented by floating point numbers, it is possible to reduce the bits used for the number representation down to single bits \cite{rastegari2016xnor}, without affecting the network accuracy.

In the context of performance evaluation of computer systems, stochastic models whose underlying stochastic processes are Markov chains,  play a key role providing a sound high-level  framework for the analysis  of software and hardware architectures.
Although the use of high-level modelling formalism greatly simplifies the specification of quantitative models (e.g., by exploiting the compositionality properties \cite{hillston:thesis}), the stochastic process underlying even a very compact model may have a number of states that makes its analysis a difficult, sometimes computationally impossible, task. In order to study models with a large state space without using approximations or resorting to simulations, one can attempt to reduce the state space of the underlying Markov chain by aggregating states with equivalent behaviours.
Lumpability is an aggregation technique 
used to cope with the state space explosion problem inherent to the computation of the stationary performance indices of large stochastic models. The lumpability method  turns out to be useful  on Markov chains exhibiting some structural regularity. Moreover, it allows one to efficiently compute the exact values of the performance indices when the model is actually lumpable. In the literature, several notions of lumping have been introduced:  ordinary and weak lumping \cite{kemeny76:basic.concepts}, exact lumping  \cite{schweitzer:aggregations}, and strict lumping~\cite{buchholz:lumpability}.

With this paper we aim to link together the work of two different communities, the first one focusing on machine learning and network compression and the second one focusing on lumping-based aggregation techniques for performance evaluation.
Even if a large number of possible efficient compression techniques has already been published, we aim instead to give a formal demonstration on how it is possible to deterministically remove some of the network parameters to obtain a smaller network with the same performance. Our method condenses many different concepts together, such as some of the ideas exploited by tensor decomposition methods, filter pruning and the lumpability used to evaluate the performance of complex systems.

The paper is structured as follows. In Section \ref{sec:related} we provide a literature review.
Section \ref{sec:prel} gives the necessary background.
Section \ref{sec:lnn} formally describes our technique  exploiting exact lumpability for quotienting NN.
Section  \ref{sec:experiments}  presents some experimental results. Finally, Section~\ref{sec:conclusion} concludes the paper.

\section{Related Work}
\label{sec:related}
To the best of our knowledge, the only paper similar to our work is \cite{prabhakar2022bisimulations}, where the authors introduce the classical notion of equivalence between systems in Process Algebra to reduce a neural network into another one semantically equivalent. They propose a filter pruning technique based on some properties of the network that does not need any data to perform the compression. They also define an approximated version of their algorithm to relax some of the strong constraints they pose on the weights of the network.

While data free pruning algorithms are convenient when a dataset is incomplete, unbalanced or missing, they usually achieve poorer results compared to data-based compression solutions.
Indeed, most pruning techniques usually require at least one stage of fine-tuning of the model. The recovery is often performed in an iterative fashion after removing a single parameter, but there are also techniques that re-train the model only after a certain level of compression has been carried out \cite{blalock2020state}. 

As defined in \cite{lin2020hrank} filter pruning techniques can be divided according to \textit{property importance} or \textit{adaptive importance}. In the first group we find pruning methods that look at intrinsic properties of the networks, and do not modify the training loss, such as \cite{prabhakar2022bisimulations,hu2016network,li2016pruning,he2019filter,ressi2022relevance,castellano1997iterative}. Adaptive importance pruning algorithms like \cite{liu2017learning,lin2019towards} usually drastically change the loss function, requiring a heavy retrain step and to look for a new proper set of hyper-parameters, despite the fact that they often achieve better performances with respect to property importance methods. 
Avoiding to re-train the network at each pruning step as in \cite{lin2020hrank,wang2021rfpruning} is usually faster than other solutions, but there is a higher risk to not being able to recover the performances.

Another option consists in deciding which parameters to remove according to the impact they have on the rest of the network \cite{molchanov2019importance,yu2018nisp}.
Finally, while most of the already mentioned methods focus on removing whole filters or kernels from convolutional layers, some other methods actually target only fully connected layers, or are made to compress classical neural networks \cite{tan2020dropnet,ashiquzzaman2019compacting}.

\section{Preliminaries}
\label{sec:prel}
In this section we formally introduce the notion of neural network in the style of \cite{prabhakar2022bisimulations}. Moreover, we recall the concept of exact lumpabibility as it has been defined in the context of continuous time Markov chains.

\subsection*{Neural Networks}
A neural network is formed by a layered set of nodes or neurons,
consisting of an input layer, an output layer and one or more hidden layers. Each node that does not belong to the input layer is annotated with a bias and an activation function. Moreover, there are weighted edges between nodes of adjacent layers. We use the following formal definition of
neural network.

For $k\in \mathbb{N}$, we denote by $[k]$ the set $\{0,1,\ldots,k\}$, by 
$(k]$ the set $\{1,\ldots,k\}$, 
by 
$[k)$ the set $\{0,\ldots,k-1\}$,
and by  
$(k)$ the set $\{1,\ldots,k-1\}$.

\begin{definition}[Neural Network]
A {\em Neural Network (NN)} is a tuple $\mathcal N =(k, \mathcal Act, \{\mathcal S_\ell\}_{\ell\in[k]}, \{ W_\ell\}_{\ell\in(k]}, \{b_\ell\}_{\ell\in(k]}, \{ A_\ell\}_{\ell\in(k]})$ where:
\begin{itemize}
    \item $k$ is the number of layers (except the input layer);
    \item $\mathcal Act$ is the set of activation functions;
    \item for $\ell\in[k]$, $\mathcal S_\ell$ is the set of nodes of layer $\ell$ with $\mathcal S_\ell \cap \mathcal S_{\ell'}=\emptyset$ for $\ell\neq \ell'$;
    \item for $\ell\in(k]$, $W_\ell: \mathcal S_{\ell-1}\times \mathcal S_\ell \rightarrow \mathbb R$ is the weight function that associates a weight with edges between nodes at layer $\ell-1$ and $\ell$;
    \item for $\ell\in(k]$, $b_\ell:\mathcal S_\ell \rightarrow \mathbb R$ is the bias function that associates a bias with nodes at layer $\ell$;
    \item for $\ell\in(k]$, $A_\ell:\mathcal S_\ell \rightarrow \mathcal Act$ is the activation association function that associates an activation function with nodes of layer $\ell$.
\end{itemize}

$\mathcal S_0$ and $\mathcal S_k$ denote the nodes in the input and output layers, respectively.

\end{definition}

In the rest of the paper we will refer to NNs in which all the activation association function are constant, i.e., all the neurons of a layer share the same activation function. Moreover, such activation functions are either ReLU (Rectified Linear Unit) or LeakyReLU, i.e., they are combinations of linear functions. So, from now on we omit the set $\mathcal Act$ from the definition of the NNs.

\begin{figure}[t]
\begin{tikzpicture}[shorten >=1pt,auto,node distance=2.3cm]
 \node[rectangle] (x1)  {$x_1$};
\node[rectangle] (x2) [right of = x1] {$x_2$};
\node[rectangle] (xm) [right of = x2] {$x_m$};
\node[node distance=0.4cm,right=of x2] {$\ldots$}; 
 \node[state,  node distance=1cm] (u1)  [below of = x1] {$u_1$};
\node[state] (u2) [right of = u1] {$u_2$};
\node[state] (um) [right of = u2] {$u_m$};
\node[rectangle, node distance=3cm] (layer1) [left of = u1] {Layer $\ell-1$};
\node[node distance=0.4cm,right=of u2] {$\ldots$}; 
\node[state] (v) [below of = u2] {$v$};
\node[node distance=2cm,right=of v] {$\ldots$}; 
\node[rectangle] (layer2) [below of = layer1] {\hskip-0.6cm Layer $\ell$ };
\node[rectangle, draw, node distance=1cm] (r) [below of = v] {$ReLU\big(\sum_{j}W_{\ell}(u_j,v) x_j + b_{\ell}(v)\big)$};
\node[state, node distance=1.5cm] (z) [below of = r] {$z$};
\node[rectangle, node distance=3cm] (layer3) [below of = layer2] {Layer $\ell+1$};
\node[node distance=2cm,right=of z] {$\ldots$}; 
\path[->] (u1) edge [swap, near start]  node {$W_{\ell}(u_1,v)$} (v);
\path[->] (u2) edge [swap, near start]  node {$W_{\ell}(u_2,v)$} (v);
\path[->] (um) edge [near start]  node {$W_{\ell}(u_m,v)$} (v);
\path[->] (v) edge  node {} (r);
\path[->] (r) edge  node {$W_{\ell+1}(v,z)$} (z);
\path[->] (x1) edge  node {} (u1);
\path[->] (x2) edge  node {} (u2);
\path[->] (xm) edge  node {} (um);

\end{tikzpicture}
    \caption{Node $v$ behaviour on input $x_1, x_2, \dots, x_m$}\label{fig:NN}
  \end{figure}
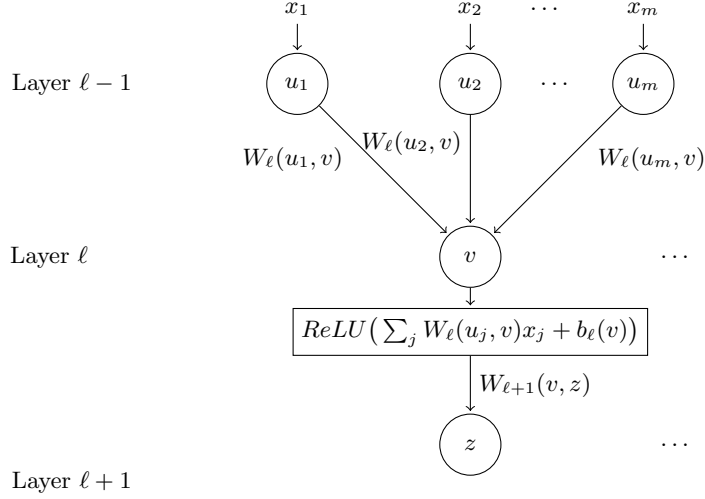 
\begin{example}
Figure \ref{fig:NN} shows the behaviour of node $v$ in Layer $\ell$. 
The input values $x_1, x_2, \dots x_m$ are propagated by nodes $u_1, u_2, \dots u_m$ respectively. 
Node $v$ computes the $ReLU$ of the weighted sum of the inputs plus the bias. 
The result of this application is the output of $v$ and it is propagated to $z$.
\end{example}

The operational semantics of a neural network is as follows. Let $v:\mathcal S_\ell\rightarrow \mathbb R$ be a valuation for the $\ell$-th layer of $\mathcal N$ and $Val(\mathcal S_\ell)$ be the set of all valuations for the $\ell$-th layer of $\mathcal N$. The operational semantics of $\mathcal N$, denoted by $[\![\mathcal N]\!]$, is defined in terms of the semantics of its layers $[\![\mathcal N]\!]_\ell$, where each $[\![\mathcal N]\!]_\ell$ associates with any valuation $v$ for layer $\ell-1$ the corresponding valuation for layer $\ell$ according to the definition of $\mathcal N$. The valuation for the output layer of $\mathcal N$ is then obtained by the composition of functions $[\![\mathcal N]\!]_\ell$.

\begin{definition}
The semantics of the $\ell$-th layer is the function $[\![\mathcal N]\!]_\ell: Val(\mathcal S_{\ell-1})\rightarrow Val(\mathcal S_\ell)$ where for all $v\in Val(\mathcal S_{\ell-1})$, $[\![\mathcal N]\!]_\ell(v)=v'$ and for all $s'\in \mathcal S_\ell$, 
$$v'(s')=A_\ell(s')\big( \sum_{s\in\mathcal S_{\ell-1}} W_\ell(s,s')v(s)+b_\ell(s')\big)\,.$$
\end{definition}

The input-output semantics of $\mathcal N$ is obtained by composing these one layer semantics. More precisely, we denote by $[\![\mathcal N]\!]^\ell$ the composition of the first $\ell$ layers so that $[\![\mathcal N]\!]^\ell(v)$ provides the valuation of the $\ell$-th layer given $v\in Val(\mathcal S_0)$ as input. Formally, $[\![\mathcal N]\!]^\ell$ is inductively defined by:
$$[\![\mathcal N]\!]^1=[\![\mathcal N]\!]_1$$
$$[\![\mathcal N]\!]^\ell= [\![\mathcal N]\!]_\ell\circ [\![\mathcal N]\!]^{\ell-1} \ \forall \ell\in(k]$$
where
$\circ$ denotes the function composition.

We are now in position to define the semantics of $\mathcal N$ as the input-output semantic function  $[\![\mathcal N]\!]$ defined below.

\begin{definition}
The input-output semantic function $[\![\mathcal N]\!]:Val(\mathcal S_0) \rightarrow Val(\mathcal S_k)$ is defined as
$$[\![\mathcal N]\!]=[\![\mathcal N]\!]^k\,.$$
\end{definition}

\subsection*{Lumpability}
The notion of \emph{lumpability} has been introduced 
in the context of performance and reliability analysis. It
provides a model aggregation technique that can be used for generating a Markov chain that is smaller than the original one while allowing one to determine exact results for the original process.

The concept of lumpability
can be formalized in terms of equivalence relations
over the state space of the Markov chain.
Any such equivalence  induces a \emph{partition} on the state space of the Markov chain
and aggregation is achieved by clustering equivalent states into macro-states, reducing the overall state space.

Let $\mathcal S$ be a finite state space. 
A (time-homogeneous) Continuous-Time Markov Chain (CTMC) over $\mathcal S$ is defined by a function 
$$Q:\mathcal S\times \mathcal S \rightarrow \mathbb R$$
such that for all $u,v\in \mathcal S$ with $u\neq v$ it holds that:
\begin{itemize}
\item $Q(u,v)\geq 0$ and 
\item   $\sum_{v\in \mathcal S, v\neq u}Q(u,v)=-Q(u,u)\,.$
\end{itemize}

A CTMC defined over $\mathcal S$ by $Q$ models a stochastic process where a transition from $u$ to $v$ can occur according to an exponential 
distribution with rate $Q(u,v)$.

Given an initial probability distribution $p$ over the states of a CTMC, one can consider the problem of computing the probability distribution to which $p$ converges when the time tends to infinity. This is the \emph{stationary} distribution and it exists only when the chain satisfies additional constraints. The stationary distribution reveals the limit behaviour of a CTMC. Many other performance indexes and temporal logic properties can be defined for studying both the transient and limit behaviour of the chain. 

Different notions of lumpability have been introduced with the aim of reducing the number of states of the chain, while preserving its behaviour \cite{inf18,buchholz:lumpability,marin:valuetools13,kemeny76:basic.concepts,formats19,acta21,schweitzer:aggregations}.
In particular, we consider here the notion of \emph{exact lumpability} \cite{buchholz:lumpability,schweitzer:aggregations}.

\begin{definition}[Exact Lumpability]
Let $(\mathcal S, Q)$ be a CTMC and $\mathcal R$ be an equivalence relation over $\mathcal S$. $\mathcal R$ is an \emph{exact lumpability} if for all $S,S'\in \mathcal R/\mathcal S$, for all $v,t\in S$ it holds that:
$$\sum_{u\in S'}Q(u,v)=\sum_{u\in S'}Q(u,t)\,.$$
\end{definition}
There exists always a unique maximum exact lumpability relation which allows to quotient the chain by taking one state for each equivalence class and replacing the rates of the incoming edges with the sum of the rates from equivalent states. 

The notion of exact lumpability is in many applicative domains too demanding, thus providing poor reductions. This issue is well-known for all lumpability notions that do not allow any form of approximation. With the aim of obtaining smaller quotients, still avoiding rough approximations, the notion of \emph{proportional lumpability} has been presented in \cite{formats19,acta21,qest21} as a relaxation of ordinary lumpability. In this paper instead we introduce to \emph{proportional exact lumpability} which is defined as follows.
\begin{definition}[Proportional Exact Lumpability]
Let $(\mathcal S, Q)$ be a CTMC and $\mathcal R$ be an equivalence relation over $\mathcal S$. $\mathcal R$ is a \emph{proportional exact lumpability} if there exists a function $\rho:\mathcal S \rightarrow \mathbb R_{>0}$ such that
for all $S,S'\in \mathcal R/\mathcal S$, for all $v,t\in S$ it holds that:
$$\rho(v)\sum_{u\in S'}Q(u,v)=\rho(t)\sum_{u\in S'}Q(u,t)\,.$$
\end{definition}
It can be proved that there exists a unique maximum proportional exact lumpability which can be computed in polynomial time. This is true also if $(\mathcal S,Q)$ is a \emph{Labelled Graph} instead of a CTMC, i.e., no constraints are imposed on $Q$. 

\begin{figure}[t]
  \begin{center}
  \resizebox{0.5\textwidth}{!}{%
\begin{tikzpicture}[shorten >=1pt,auto,node distance=2cm]
  \node[state,fill=white!70!green] (3)                {3};
 \node[state,fill=white!70!magenta] (1) [left of = 3] {1};
\node[state,fill=white!70!green] (2) [above of = 3] {2};
\node[state,fill=white!70!green] (4) [below of = 3] {4};
\node[state,fill=white!70!yellow] (5) [right of = 2] {5};
\node[state,fill=white!70!yellow] (6) [right of = 3] {6};
\node[state,fill=white!70!yellow] (7) [right of = 4] {7};
\node[state,fill=white!70!cyan] (8) [right of = 6] {8};
\path[->] (1) edge [bend left=20]  node {$2$} (2)
(2) edge   node {$3$} (5)
(5) edge [swap, bend left=20]   node {$1$} (8);
\path[->] (1) edge   node {$1$} (3)
(3) edge [swap, bend left=20, near start]   node {$3$} (5);
\path[->] (1) edge [bend right=20]  node {$2$} (4)
(4) edge  [swap] node {2} (7)
(7) edge [bend right=20]   node {1} (8)
(8) edge [swap, bend right=90, looseness=1.5]  node {$4$} (1);
\path[->] (4) edge [swap, bend right=20, near end]  node {1} (6)
(6) edge   node {5} (8);
\path[->] (3) edge [bend right=20, near start]  node {4} (7);
   \path[->] (2) edge [bend left=20, near end]  node {1} (6);
\end{tikzpicture}}
    \end{center}
    \caption{Proportionally exact lumpable CTMC.}\label{fig:pel}
  \end{figure}
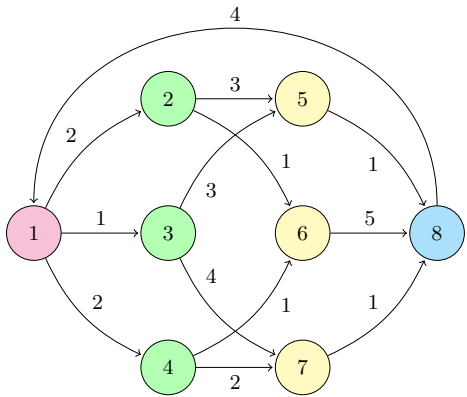 
  
\begin{example}
Figure \ref{fig:pel} shows a proportionally exact lumpable Markov chain with respect to the function $\rho$ defined as: $\rho(1)=1, \rho(2)=1, \rho(3)=2, \rho(4)=1,\rho(5)=1, \rho(6)=2, \rho(7)=1, \rho(8)=1$ and the equivalence classes $S_1=\{1\}, S_2=\{2,3,4\}, S_3=\{5,6,7\}, S_4=\{8\}$.

\end{example}

\section{Lumping Neural Networks}
\label{sec:lnn}
The idea of exploiting exact lumpability for quotienting NN has been proposed in \cite{prabhakar2022bisimulations} where a notion of pre-sum preserving backward bisimulation has been considered. It can be easily observed that such a notion coincides with that of exact lumpability. The term (probabilistic) bisimulation is standard in the area of Model Checking, where (probabilistic) temporal logical properties are used for both specifying and synthesizing systems having a desired behaviour (see, e.g., \cite{dang2015parameter}). Since such logics usually formalize the behaviours in terms of forward temporal operators, the bisimulation notions tend to preserve the rates of the outgoing edges \cite{sproston2004backward}. However, as proved in \cite{prabhakar2022bisimulations}, in order to preserve the behaviour of a NN it is necessary to refer to the rates/weights of the incoming edges. This is referred to as \textit{backward probabilistic bisimulation} and coincides with the well-known notion of \textit{exact lumpability} used in the area of performances evaluation. 

In this paper we extend the proposal of \cite{prabhakar2022bisimulations}. We prove that in the case of ReLU and LeakyReLU activations, proportional exact lumpability preserves the behaviour of the network allowing to obtain smaller quotients. It does not require any retraining step and it ensures the same behaviour on all possible inputs. Moreover, since the neural networks we refer to are acyclic it can be computed in linear time. 

\begin{definition}[Proportional Exact Lumpability over a NN]\label{def1}
Let $\mathcal N$ be a NN. 
Let $\mathcal R=\cup_{\ell\in [k)}\mathcal R_\ell$ be 
such that $\mathcal R_\ell$ is an equivalence relation over $\mathcal S_\ell$, for all $\ell\in (k)$ and $\mathcal R_0$ is the identity relation over $\mathcal S_0$. We say that $\mathcal R$ is a \emph{proportional exact lumpability} over $\mathcal N$ if for each $\ell\in (k)$ there exists $\rho_\ell:\mathcal S_\ell\rightarrow \mathcal R_{>0}$ such that for all $S\in \mathcal S_\ell$, for all $S'\in \mathcal S_{\ell-1}$, for all $v,t\in S$ it holds that:
$$\begin{array}{c}
 \rho_\ell(v)b_\ell(v)= \rho_\ell(t)b_\ell(t)\, ,\\[0.2cm]
\rho_\ell(v)\sum_{u\in S'}W_\ell(u,v)=\rho_\ell(t)\sum_{u\in S'}W_\ell(u,t)\,.
\end{array}$$
\end{definition}
There are some differences with respect to the definition of proportional exact lumpability over CTMCs. First, we impose that two equivalent neurons have to belong to the same layer. However, we could have omitted such restriction from the definition and proved that neurons from different layers are never equivalent. This is an immediate consequence of the fact that we refer to acyclic NNs. Moreover, we demand that on input and output nodes the only admissible relation is the identity. This is a substantial difference. Since the nodes in the input layer have no incoming edges the definition of proportional lumpability given over CTMCs allows to collapse them. However, the input nodes in NNs hold the input values that have to be propagated, so they cannot be collapsed. This is true also for the output nodes, since they represent the result of the computation.

It can be proved that there always exists a unique maximum proportional exact lumpability over a NN.
If we use proportional exact lumpability for reducing the dimension of a NN by collapsing the equivalent neurons, we have to modify the topology and the weights of the NN as formalized below.
\begin{definition}[Proportional Reduced NN]
Let $\mathcal N=(k,\{\mathcal S_\ell\}_{\ell\in[k]},$ $\{ W_\ell\}_{\ell\in(k]},$ $\{b_\ell\}_{\ell\in(k]}, \{ A_\ell\}_{\ell\in(k]})$ be a NN. Let $\mathcal R$ be a proportional exact lumpability over $\mathcal N$.
The NN $\mathcal N/\mathcal R= (k, \{\mathcal S'_\ell\}_{\ell\in[k]}, \{ W'_\ell\}_{\ell\in(k]}, \{b'_\ell\}_{\ell\in(k]}, \{ A'_\ell\}_{\ell\in(k]})$ is defined by:
\begin{itemize}
    \item $\mathcal S'_\ell=\{[v]\:|\: [v]\in \mathcal S_\ell/\mathcal R\}$, where $v$ is an arbitrarily chosen representative for the class;
    \item $W_{\ell}'([u],[v])=\rho_{\ell-1}(u)\sum_{w\in[u]}\frac{W_{\ell}(w,v)}{\rho_{\ell-1}(w)} $;
    \item $b_\ell'([v])=b_\ell(v)$;
    \item $A'_\ell([v])=A_\ell(v)$.
\end{itemize}
\end{definition}
Despite the arbitrary choice of the representative, we can prove that the reduced NN's behaviour coincides with that of the initial one over all the inputs.
\begin{theorem}\label{maintheo}
Let $\mathcal N$ be a NN and $\mathcal R$ be a proportional exact lumpability over $\mathcal N$. It holds that
$$[\![\mathcal N/\mathcal R]\!]=[\![\mathcal N]\!]\,.$$
\end{theorem}
\begin{proof}
Sketch.
Let us focus on two neurons $v$ and $t$ belonging to layer $1$ that are equivalent in $\mathcal R_1$. Let $ReLU$ be the activation function for both of them.

On input $x_1,x_2,\dots x_m$ for the nodes $u_1,u_2,\dots,u_m$ of layer $0$ the nodes $v$ and $t$ take values
$Val(v)=ReLU(\sum_{j=1}^m W_1(u_j,v)x_j+b_1(v))$ and
$Val(t)=ReLU(\sum_{j=1}^m W_1(u_j,t)x_j+b_1(t))$, respectively.
However, since $v$ and $t$ are equivalent, it holds that:
$$\sum_{j=1}^m W_1(u_j,t)x_j+b_1(t)=\frac{\rho_1(v)}{\rho_1(t)}\sum_{j=1}^m W_1(u_j,v)x_j+b_1(v)$$
Since $\rho_1(v)$ and $\rho_1(t)$ are positive numbers, we get that:
$$\begin{array}{lcl}
Val(t) &= &ReLU(\sum_{j=1}^m W_1(u_j,t)x_j+b_1(t))\\& = &\frac{\rho_1(v)}{\rho_1(t)}ReLU(\sum_{j=1}^m W_1(u_j,v)x_j+b_1(v))=\frac{\rho_1(v)}{\rho_1(t)}Val(v)\,.
\end{array}$$
Let now $z$ be a neuron of layer $2$. The value of $z$ depends on $$W_2(v,z)Val(v)+W_2(t,z)Val(t)=(W_2(v,z)+\frac{\rho_1(v)}{\rho_1(t)}W_2(t,z))Val(v)$$ 
So, the definition of $W_2'$ takes care of the fact that in the reduced network $v$ represents the equivalence class, while $t$ has been ``eliminated''. Such definition ensures that the value of neuron $z$ is unchanged. 

A formal proof can be obtained generalizing the above arguments.
\qed
\end{proof}

\begin{figure}[t]
  \begin{center}
  \resizebox{1.05\textwidth}{!}{%
\begin{tikzpicture}[shorten >=1pt,auto,node distance=2.3cm]
 \node[state,  node distance=1cm] (u1)  [below of = x1] {$u_1$};
\node[state] (u2) [right of = u1] {$u_2$};
\node[state] (um) [right of = u2] {$u_m$};
\node[rectangle, node distance=3cm] (layer1) [left of = u1] {Layer $\ell-1$};
\node[node distance=0.4cm,right=of u2] {$\ldots$}; 
\node[state] (v) [below of = u1] {$v$};
\node[state,red] (t) [below of = um] {$t$};
\node[draw, cross out,red] (x) [below of = um] {$t$};
\node[node distance=2cm,right=of t] {$\ldots$}; 
\node[rectangle] (layer2) [below of = layer1] {\hskip-0.6cm Layer $\ell$ };
\node[state, node distance=4cm] (z) [below of = u2] {$z$};
\node[rectangle, node distance=1.8cm] (layer3) [below of = layer2] {Layer $\ell+1$};
\node[node distance=2cm,right=of um] {$\ldots$}; 
\node[node distance=2cm,right=of z] {$\ldots$}; 
\path[->] (u1) edge [swap, near end]  node[xshift=1mm] {\tiny $W_{\ell}(u_1,v)$} (v);
\path[->] (u2) edge [swap,  near start] node[yshift=-4mm, xshift=-1mm] {\tiny  $W_{\ell}(u_2,v)$} (v);
\path[->] (um) edge [near end]  node[yshift=-2mm, xshift=-6mm] {\tiny $W_{\ell}(u_m,v)$} (v);
\path[->] (v) edge  [swap] node {\tiny $W_{\ell+1}(v,z)+{\color{black!40!green}{W_{\ell+1}(t,z)/\rho}}$} (z);
\path[->,red] (u1) edge [swap, near end]  node[draw, strike out, yshift=3mm, xshift=-2mm] {} node[yshift=-2mm, xshift=6mm]  {\tiny $\rho W_{\ell}(u_1,v)$} (t);
\path[->,red] (u2) edge [near start]  node[draw, strike out, yshift=-8mm, xshift=6mm] {}  node[yshift=-4mm, xshift=2mm]  {\tiny $\rho W_{\ell}(u_2,v) $ } (t);
\path[->,red] (um)  edge [near end]  node[draw, strike out, yshift=3mm, xshift=-1mm] {} node {\tiny $\rho W_{\ell}(u_m,v)$}  (t);
\path[->,red] (t) edge  node[draw, strike out, rotate=70, yshift=3mm, xshift=-4mm] {} node {\tiny $W_{\ell +1}(t,z)$} (z);
\end{tikzpicture}}
  \end{center}
    \caption{Pruning one node and updating the network.}\label{fig:NNreduced}
  \end{figure}

\begin{example}
    Figure \ref{fig:NNreduced} shows how the pruning technique works on two nodes $v, t$. 
    In particular, $t$ input weights are proportionals to $v$'s. 
    The algorithm proceeds in two steps. Firstly, $t$ is deleted together with all its input and output edges. Secondly, the weight from $v$ to $z$ is modified by adding $W_{\ell + 1}(t, z) / \rho$.
\end{example}

The maximum proportional exact lumpability over $\mathcal N$ together with the reduced network can be efficiently computed by proceeding top-down from layer $1$ to $k-1$. Since the network is acyclic, each layer is influenced only by the previous one. Hence, the computation is linear with respect to the number of edges of the network.
\begin{theorem}
Let $\mathcal N$ be a NN. There exists a unique maximum proportional exact lumpability $\mathcal R$ over $\mathcal N$.
Moreover, $\mathcal R$ and $\mathcal N/\mathcal R$ can be computed in linear time with respect to the size of $\mathcal N$, i.e., in time $\Theta(\sum_{\ell\in (k]}|\mathcal S_{\ell-1}\times \mathcal S_{\ell}|)$.
\end{theorem}

Intuitively, Theorem \ref{maintheo} exploits the following property of ReLU (LeakyReLU):
$$\forall y \in \mathbb R \:\forall r\in \mathbb R_{>0} \: \:ReLU(r*y)=r*ReLU(y)\,.$$
This allows us to remove some neurons exploiting the proportionality relation with others. In order to guarantee the correctness of the removal on all possible inputs, as stated in Theorem \ref{maintheo}, it is not possible to exploit less restrictive relationships than proportionality. This fact can also be formally proved, under the hypothesis that the input set is sufficiently rich.
However, one could ask what happens if we move from a simple proportionality relation to a linear dependence. For instance, what happens if in Definition \ref{def1} we relax the two equations by considering that $t$ is a linear combination of $v_1$ and $v_2$, i.e.:
$$\begin{array}{c}
 \rho_\ell(t)b_\ell(t) =\rho_\ell(v_1)b_\ell(v_1)+\rho_\ell(v_2)b_\ell(v_2)\,,\\[0.2cm]
\rho_\ell(t)\sum_{u\in S'}W_\ell(u,t)=\rho_\ell(v_1)\sum_{u\in S'}W_\ell(u,v_1) + \rho_\ell(v_2)\sum_{u\in S'}W_\ell(u,v_2)\,.
\end{array}$$
In this case we could eliminate $t$ by including its contribution on the outgoing edges of both $v_1$ and $v_2$. Unfortunately, the behaviour of the network is preserved only for those input values $x_1,x_2,\dots,x_m$ which ensure that 
$\sum_{j=1}^m W_\ell(u_j,v_1)x_j+b_\ell(v_1)$ and $\sum_{j=1}^m W_\ell(u_j,v_2)x_j+b_\ell(v_2)$ have the same sign, since 
$$\forall y_1,y_2 \in \mathbb R,\ \: \forall r_1,r_2\in \mathbb R_{>0},$$ $$ReLU(r_1*y_1+r_2*y_2)=r_1*ReLU(y_1)+r_2*ReLU(y_2) \mbox{ iff }y_1*y_2\geq 0\,.$$

In other terms our analysis points out that reduction techniques based on linear combinations of neurons 
can be exploited without retraining the network only when strong hypothesis on the sign of the neurons hold. 

More sophisticated methods that 
 exploit \emph{Principal Component Analysis} can be seen as a further shift versus approximation, since they do not only involve linear combinations of neurons, but also a base change and the elimination of the less significant dimensions.

\section{Experimental Results}
\label{sec:experiments}
To assess the robustness of our method we set up some simple experiments where we implemented the neural network pruning by lumping. In particular, we want to show how the accuracy is affected when the weights of the node to prune are not simply proportional to the weights of another node in the same layer, but they are instead a linear combination of the weights of two or more other nodes.

We designed and trained a simple \textit{Convolutional Neural Network(CNN)} made of two convolutional blocks (32 3 $\times$ 3 filters each, both followed by a maxpooling layer) and after a simple flatten we add three fully connected layers (fc), with 16, 128 and 10 nodes each, where the last one is the softmax layer. As required by our method, we use only ReLU activations, except for the output layer. We used the benchmark MNIST dataset, consisting of 7000 28$\times$28 greyscale images of handwritten digits divided into 10 classes.

After a fast training of the model we focused on the second last fully connected layer for our pruning method. We randomly selected a subset of nodes in this layer and then manually overwrote the weights of the rest of the nodes in the same layer as linear combinations of the fixed ones. We then froze this synthetic layer and retrained the network to recover the lost accuracy. The resulting model presents a fully connected layer with 2176 (2048 weight + 128 bias) parameters that can be the target of our pruning method. 

\begin{figure}
	\centering
	\vspace{-1em}
	\includegraphics[scale=0.68]{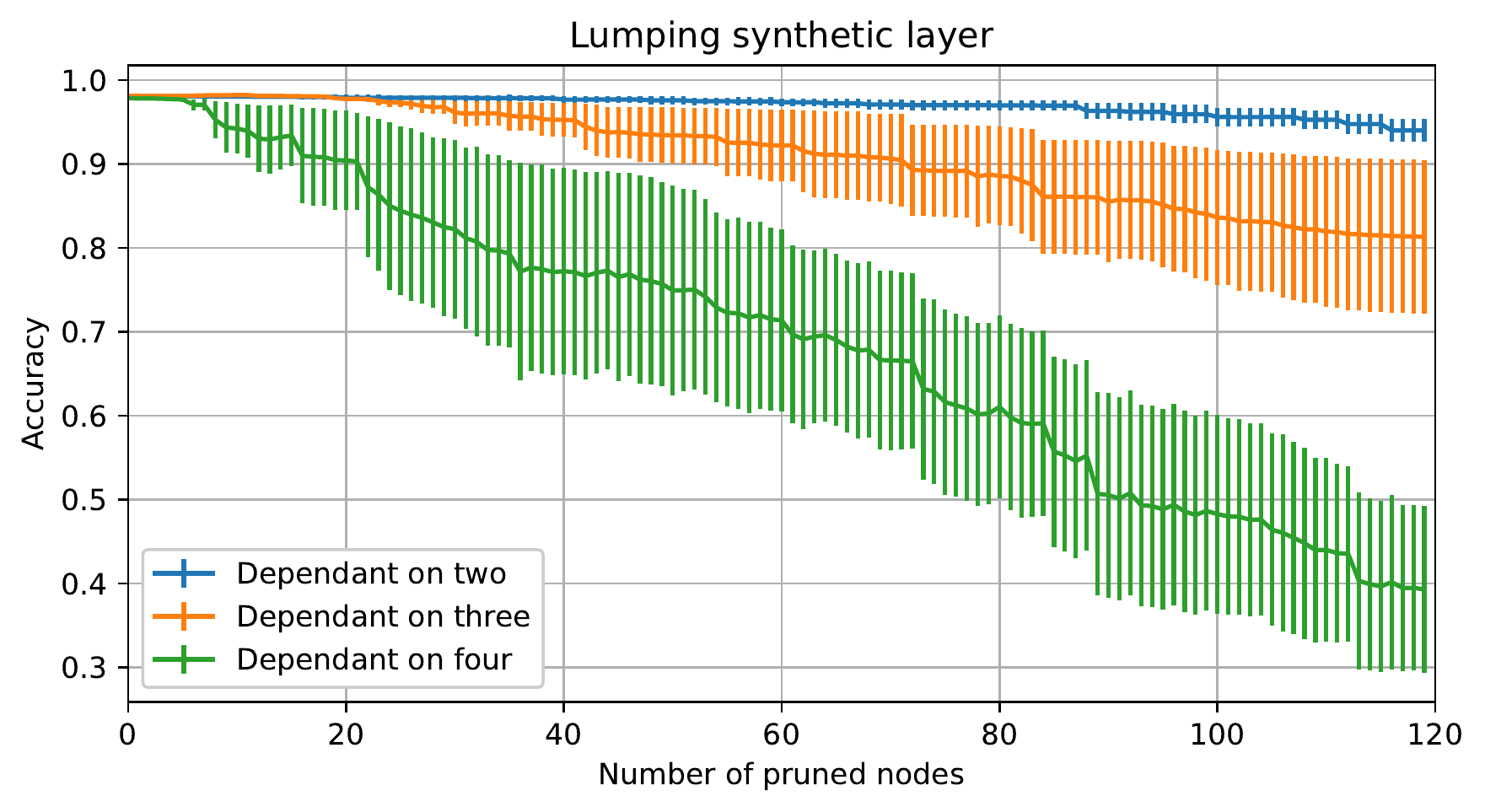}
	\caption{Accuracy loss when pruning nodes which incoming weights are linear combination of two, three and four other nodes' weights in the same layer.}
	\label{fig:Exp}
	\vspace{-2em}
\end{figure}

During the first round of experiments we confirmed that if the weights in the fixed subset have all the same sign, then our method prunes the linearly dependant vectors and the updating step does not introduce any performance~loss. Differently,
as illustrated in Figure \ref{fig:Exp}, when the weights in the subset have different sign, the updating step can introduce some loss. This happens only  in the case that the weights are a linear combination of two or more of the weights incoming to the other nodes in the synthetic layer. In particular, the accuracy drops faster as the number of nodes involved in the linear combination increases.

\section{Conclusion}
\label{sec:conclusion}
In this paper we present a data free filter pruning compression method based on the notion of lumpability. Even though we impose rigid constraints on the weights in order to obtain a reduced network, in doing so we also demonstrate how the resulting model exhibits the same exact behaviour. Regardless the limitations of our method, this work opens the door to a new research field where the aggregation techniques typical of performance evaluation  are adopted in network compression, usually explored only by the machine learning community. In the future, we would like to further analyze how our algorithm works for different study cases, and in particular to test how an approximation of the linear dependence would affect the accuracy under different conditions. Another interesting experiment would be to use SVD on the fully connected layers to estimate how many vectors are linearly independent and therefore compute the reduction potentially achieved by our method, especially for quantized networks.

\subsubsection*{Acknowledgements.}
This work has been partially supported by the Project PRIN 2020  “Nirvana - Noninterference and Reversibility Analysis in Private Blockchains” and by the Project GNCS 2022 “Propriet\`a qualitative e quantitative di sistemi reversibili”.

\bibliographystyle{splncs04}
\bibliography{main}

\end{document}